\documentclass[11pt,a4paper]{article}
\usepackage[hyperref]{moe-arxiv}
\usepackage{times}
\usepackage{latexsym}

\usepackage{listings}
\usepackage{xcolor}

\definecolor{codegreen}{rgb}{0,0.6,0}
\definecolor{codegray}{rgb}{0.5,0.5,0.5}
\definecolor{codepurple}{rgb}{0.58,0,0.82}
\definecolor{backcolour}{rgb}{0.95,0.95,0.92}

\lstdefinestyle{mystyle}{
    backgroundcolor=\color{backcolour},   
    commentstyle=\color{codegreen},
    keywordstyle=\color{magenta},
    numberstyle=\tiny\color{codegray},
    stringstyle=\color{codepurple},
    basicstyle=\ttfamily\footnotesize,
    breakatwhitespace=false,         
    breaklines=true,                 
    captionpos=b,                    
    keepspaces=true,                 
    numbers=left,                    
    numbersep=5pt,                  
    showspaces=false,                
    showstringspaces=false,
    showtabs=false,                  
    tabsize=2
}

\usepackage{microtype}
\usepackage{graphicx}
\usepackage{subcaption}
\usepackage{array}
\usepackage{lettrine}

\usepackage{amsmath}
\usepackage{booktabs}
\usepackage{multirow}
\usepackage{svg}

\aclfinalcopy % Uncomment this line for the final submission
\title{Scalable and Efficient MoE Training for Multitask Multilingual Models}

\author{Young Jin Kim$^{\ast}$,
  Ammar Ahmad Awan$^{\ast}$ \\
  {\bf Alexandre Muzio,
  Andres Felipe Cruz Salinas,
  Liyang Lu,
  Amr Hendy } \\
  {\bf Samyam Rajbhandari,
  Yuxiong He,
  Hany Hassan Awadalla} \\
  Microsoft, One Microsoft Way, Redmond, WA 98052, USA \\
  \texttt{\small\{youki,amawa,alferre,ancruzsa,liylu,amrhendy,samyamr,yuxhe,hanyh\}@microsoft.com}
}

\date{}

\begin{document}
\maketitle
\begin{abstract}

The Mixture of Experts (MoE) models are an emerging class of sparsely activated deep learning models that have sublinear compute costs with respect to their parameters.  In contrast with dense models, the sparse architecture of MoE offers opportunities for drastically growing model size with significant accuracy gain while consuming much lower compute budget.  However, supporting large scale MoE training also has its own set of system and modeling challenges.

To overcome the challenges and embrace the opportunities of MoE, we first develop a system capable of scaling MoE models efficiently to trillions of parameters.  It combines multi-dimensional parallelism and heterogeneous memory technologies harmoniously with MoE to empower 8x larger models on the same hardware compared with existing work.
Besides boosting system efficiency, we also present new training methods to improve MoE sample efficiency and leverage expert pruning strategy to improve inference time efficiency. 
By combining the efficient system and training methods, we are able to significantly scale up large multitask multilingual models for language generation which results in a great improvement in model accuracy.
A model trained with 10 billion parameters on 50 languages can achieve state-of-the-art performance in Machine Translation (MT)   and multilingual natural language generation tasks.

The system support of efficient MoE training has been implemented and open-sourced with the DeepSpeed\footnote{\url{https://github.com/microsoft/DeepSpeed}} library.

\end{abstract}

\section{Introduction}
\label{sec:intro}
\renewcommand{\thefootnote}{$\ast$} 
\footnotetext{First authors with equal contribution.}
\renewcommand{\thefootnote}{\arabic{footnote}}
It has been a well-known fact that increasing the model size results in accuracy improvements for various machine learning models \cite{openai_scaling}. As a result, pretrained language models have grown in size from a few hundred million parameters (BERT \cite{devlin2018bert})  to hundreds of billions of parameters (GPT-3 \cite{gpt-3}). Existing research shows that we have not reached a saturation point, and this trend of increasing model size is expected to continue.

However, sustaining this growth requires tremendous compute resources, requiring millions of GPU compute hours. For example, the GPT-3 175B parameter model takes 168 days to train on 256 NVIDIA A100 GPUs based on the latest optimizations \cite{narayanan2021efficient}, while a model with a trillion parameters will take over a year even with 512 NVIDIA A100 GPUs, making these models prohibitively expensive. 

From a model architecture standpoint, previous works show that combining Multitask Learning (MTL) \cite{caruana1997multitask} and multilingual learning can significantly improve training for large scale pretrained language models \cite{aharoni-etal-2019-massively}, \cite{xue-etal-2021-mt5} and \cite{ wang-etal-2020-multi}. Such multitask multilingual learning paradigm is leveraging the inductive bias and regularization from several tasks and languages simultaneously to perform better on various downstream tasks. However, it is not easy to scale up such complex multitask multilingual models. 
The Mixture of Experts (MoE) architecture has been successfully utilized to scale massive large scale multilingual models \cite{lepikhin2020gshard} and \cite{fedus2021switch}. While multilingual models are considered one dimension of multitask training; we extend this by considering models that are trained on multilingual and multitasks simultaneously to perform better on various downstream tasks. This multi-dimensional multitask training requires model scaling capabilities along with better regularization and inductive  bias.

MoE models are a natural match for this setup, since their inductive bias allows learning a generalized model from various tasks while providing the necessary specialization for multitask and multilingual training through sparse experts. MoE exhibits unique opportunities in multitask and multilingual learning. 

MoE models also provide a potential solution to this prohibitive cost of training dense models. 
MoE incurs only sublinear compute costs with respect to the model size by sparsely activating models that selectively utilize a \emph{subset} of the model parameters for any given input. 
For example, the cost of training the Switch Transformer \cite{fedus2021switch} with 1.6 trillion parameters is less than the compute budget required to train a 10 billion parameter dense model. MoE offers the benefits of a larger model at a constant compute cost, making it an attractive solution towards improving model accuracy without paying the compute premium of massive dense models.
However, large scale MoE models bring forward their own set of unique challenges for system design as well as efficient training and inference methods. 

\subsection{System challenges} 

In terms of model scale, the number of parameters in an MoE model is determined by two factors: i) the size of the base model, which is the number of parameters in the model without any expert layers, and ii) the number of experts in the model. Increasing the former increases both the model size and computation cost, while increasing the latter increases the model size but not the compute cost. 

While increasing experts does not increase the compute cost, there is a limit to the achievable quality improvement from increasing just the number of experts \cite{fedus2021switch}. To achieve a target accuracy, while minimizing the computation cost, a balance must be maintained between the base model size and the number of experts. Therefore, effectively scaling an MoE model requires scaling both the base model and the number of experts.

However, doing so is challenging. Techniques like tensor-slicing \cite{shoeybi2020megatronlm} and pipeline parallelism \cite{harlap2018pipedream} designed to scale dense models are not effective in scaling the number of experts due to the sparse nature of MoE models. On the other hand, expert parallelism~\cite{fedus2021switch} designed to scale the model size by increasing the number of experts does not apply to non-expert parameters by definition. Creating a scalable MoE system thus requires combining parallelism techniques targeting both expert and non-expert parameters of a model.

Existing MoE systems like GShard~\cite{lepikhin2020gshard} and Switch~\cite{fedus2021switch} use a combination of expert, data, and model parallelism or a subset of them. However, this approach has two major limitations: i) They replicate the base model (part of the model without expert parameters) across data-parallel GPUs, resulting in wasted memory, and ii) they require model parallelism to scale the base model to go beyond 1.4 billion parameters, requiring nontrivial model code refactoring. 

In addition to these limitations, all existing MoE systems are also fundamentally limited by the device memory wall. This is in stark contrast to dense models.  
For dense models, the largest model size that can be trained is limited by the time it takes to train the model on today's compute clusters. A trillion parameter model can take two years to train even with 512 A100 GPUs. On the contrary, due to the sublinear computation cost of MoE models with respect to their parameters, even a multi-trillion parameter model can be trained within a few weeks on the same cluster. However, such a model simply does not fit in the device memory of 512 A100 GPUs. Therefore, regardless of the choice of parallelism, MoE model scale is fundamentally limited by the available device memory.

\subsection{Training and inference challenges}
In terms of training and inference methods to deal with MoE models, we note some important challenges and open questions.

First, MoE models have an intrinsic problem of limited capacity and imbalanced expert utilization. Even with multiple methods such as adding auxiliary balancing loss and noised gating algorithms presented by previous works~\cite{lepikhin2020gshard, fedus2021switch}, it is challenging to handle overflown tokens with a limited capacity of experts. This can be mitigated by using higher number of capacity factor, but this results in inefficient training by adding redundant computation and communication across GPUs.

Second, by increasing the number of model parameters, it is important for the model to not overfit but learn more general representations with a fixed amount of data. Though many regularization techniques have been developed over the years, MoE architecture provides a good opportunity to regularize the model by exploiting the architecture's intrinsic characteristics.
In addition, as the number of model parameters increases, it has been shown that the initialization of model parameters is important to achieve more sample efficient and stable training ~\cite{fedus2021switch}. However, it has not been well studied so far.

In terms of model deployments, even with the same amount of forward computation, increased model size requires more GPU memory which results in inefficiency at runtime. Here, expert layers in MoE models naturally learn a certain level of autonomy and specialization by activating only one or two experts at a time. This brings interesting opportunities to explore better model methods for model initialization and runtime efficiency. 

\subsection{Proposed Solution} 

To overcome the challenges and embrace the opportunities of MoE, we develop:

1) DeepSpeed MoE, a system capable of scaling MoE models to trillions of parameters through a seamless orchestration between five different forms of parallelism targeting both expert and non-expert parameters, while at the same time leveraging CPU memory to go beyond the GPU memory wall (Section \ref{sec:ds_moe}). 

2) Effective training methods for token selection which handles overflown tokens more efficiently and improves the sample efficiency and two ways of utilizing experts as composable elements, which are aggregating experts for training efficiency and experts pruning strategy to improve inference time (Section \ref{efficient_moe}).

3) Effective training recipes to scale up multitask and multilingual language models with MoE model architecture. We show how MoE models are able to efficiently scale up while learning a more generalized  model from the multitask training that outperforms previous models on several downstream tasks. (Section \ref{sec:model_train}).

We utilize the proposed DeepSpeed MoE system and effective training methods and recipes to train a family of highly efficient large scale language models called Z-code M3 (Multilingual Multitask MoE). Z-code M3 has been trained as encoder-decoder model with 10 billion parameters on 50 languages and multiple pretraining tasks. We explore the model's performance on various downstream tasks such as machine translation and cross-lingual summarization. Details about the model architecture are presented at length in Section \ref{sec:model_train} while several state-of-the-results are presented for Z-code M3 in Section~\ref{sec:exp}.

\section{DeepSpeed MoE: A scalable system for training MoE models}
\label{sec:ds_moe}

In this section, we take a deeper look at the technology inside DeepSpeed MoE. We then discuss the efficiency and scalability of DeepSpeed MoE. We conclude this section with a brief overview of the DeepSpeed MoE API.

\subsection{Seamless orchestration of parallelism}

DeepSpeed MoE supports five different forms of parallelism shown in Table \ref{tab:multi-dimension}. These techniques have been discussed in the literature~\cite{lepikhin2020gshard, rajbhandari2020zero}. We provide a brief overview on these techniques and how they work together in DeepSpeed MoE to scale both expert and non-expert parameters beyond what is possible in existing MoE Systems.

\begin{table*}[htbp]
\begin{center}\small
{
\scriptsize
\begin{tabular}{ll} \toprule 
Parallelism dimensions & Benefit \\ \midrule
Expert & Scales the model size by increasing the number of experts \\
Expert + data & Accelerates training throughput by scaling to multiple data parallel groups \\
Expert + ZeRO	& Partitions the nonexpert parameters to support larger base models \\
Expert + data + model (tensor-slicing) &	Supports massive hidden sizes and even larger base models than Expert + ZeRO \\
Expert + ZeRO + model (tensor-slicing) &	Supports massive hidden sizes and even larger base models than Expert + ZeRO \\
Expert + ZeRO-Offload + model (tensor slicing) & Leverages both GPU and CPU memory for large MoE models on limited GPU resources \\
\bottomrule
\end{tabular}
}
\end{center}
\caption{Flexible parallelism dimensions supported by DeepSpeed MoE.}
\label{tab:multi-dimension} 
\end{table*}

\textbf{Data Parallelism (DP):} For a model that fits in the GPU memory for training, data parallelism (DP) is used to scale training to multiple devices. In DP, model parameters are replicated on each device. At each step, a mini-batch is divided evenly across all the data parallel processes, such that each process executes the forward and backward propagation on a different subset of data samples, and uses averaged gradients across processes to update the model locally.

\textbf{Tensor Slicing: } When a model does not fit in the device memory, various techniques can be used to split the model among different GPUs (or devices). This is loosely called model parallelism (MP). One such technique is called tensor-slicing \cite{shoeybi2020megatronlm, shazeer2018mesh} where individual tensors such as parameters or activations can be split (or sliced) across multiple devices to reduce memory requirement per GPU. Tensor slicing allows going beyond the memory limitation of a single GPU. 

\textbf{Expert Parallelism: } Expert parallelism ~\cite{fedus2021switch} is another form of model parallelism targeting experts parameters of an MoE models. In expert parallelism, different experts are placed on different devices, and executed in parallel. When experts reside on different GPU devices, explicit communication using All-to-All primitive is necessary.

Existing MoE systems offer a combination of expert, model and data parallelism, where expert parallelism allows scaling the model size by increasing the number of experts, model parallelism allows for scaling the non-expert parameters, and data parallelism allows scaling across a large number of GPUs. However, as discussed in Sec.~\ref{sec:intro}, this approach replicates model states across data parallel devices resulting in i) memory redundancy across DP and ii) requiring model parallelism to increase non-expert parameters which needs major code refactoring.

\textbf{Zero Redundancy Optimizer (ZeRO): } To address this problem, DeepSpeed MoE supports ZeRO parallelism in addition to expert, model and data parallelism. ZeRO \cite{rajbhandari2020zero} is a model efficient variation of data parallelism. ZeRO partitions the model states such as optimizer states, parameters and gradients during training instead of replicating them. There are three stages of ZeRO corresponding to the partitioning of each of the aforementioned model states. For the purpose of this paper, ZeRO implies ZeRO Stage 2, that partitions optimizer states and gradients.   

DeepSpeed MoE can seamlessly replace data parallelism with ZeRO which removes the optimizer state and gradient redundancy across data parallel processes. For large transformer models, optimizer states and gradients can consume over $85\%$ of the total memory during training. Partitioning these across GPUs instead of replicating them, free up memory to allow for significant increase in non-expert parameters. 

For example, through ZeRO, DeepSpeed MoE can support up to 15 billion non-expert parameters \footnote{The base model size (non-expert parameters) of 15 billion parameters is calculated based on the NVIDIA A100 40GB device cluster.}, larger than the non-expert parameters of the Switch Transformer (Switch-C 1.6 trillion parameters with a base model size of less than 5 billion parameters). This is over 10x larger compared with existing MoE systems that can support only 1.4 billion non-expert parameters without adding the complexity of model parallelism. When combined with model parallelism, the base model (non-expert part of the MoE model) alone can have over 100 billion parameters, which is simply not possible with existing systems.

\subsection{Transcending the GPU-memory-wall with ZeRO-Offload}

While the support for ZeRO enables DeepSpeed MoE to scale both the expert and non-expert parameters, the largest MoE model that can be trained on a GPU cluster is still limited by the aggregate device memory. To go beyond the GPU memory, DeepSpeed MoE supports ZeRO-Offload~\cite{ren2021zero-offload} in addition to ZeRO. ZeRO-Offload is an extension of ZeRO, where the optimizer states and gradients are not only partitioned across different devices, but also offloaded to the CPU memory. 

\begin{figure}[htbp]
 \centering
\includegraphics[width=1.0\linewidth]{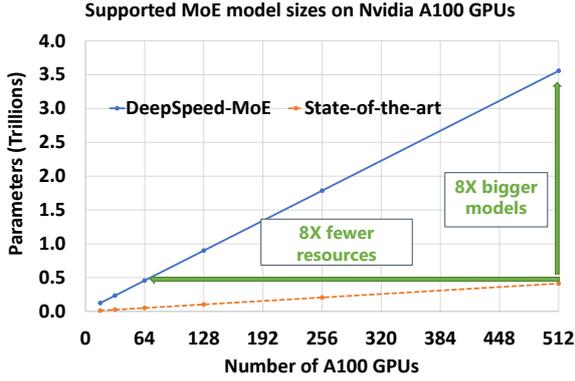}
\caption{DeepSpeed MoE scaling.}
\label{fig:dsscale}
\end{figure}

As mentioned earlier, for large transformer models, optimizer states and gradients account for over $85\%$ of the total memory consumption. Therefore, offloading them to the CPU memory significantly reduces the aggregate GPU memory consumption allowing DeepSpeed MoE to unlock unprecedented model scale. For example, it supports MoE models with over 3.5 trillion parameters on 512 NVIDIA A100 GPUs. This is an 8x increase in the total model size (3.5 trillion vs. 400 billion) compared with existing MoE systems that are limited by the total GPU memory. Alternatively, DeepSpeed MoE achieves the same model scale with 8x fewer resources (400 billion on 64 GPUs instead of 512 GPUs), as shown in Figure \ref {fig:dsscale}.

\subsection{Efficiency and Multi-GPU Scalability}

\textbf{Efficiency} DeepSpeed MoE achieves excellent computational efficiency. For transformer MoE with a hidden size of 4,096 and batch size per expert of 8,192 tokens, DeepSpeed MoE achieves over 120 TFLOPS in performance as shown in Table \ref{tab:flops}. %This is on par with dense model training scenarios. 

In general, we observe that for a transformer based MoE model, the computational efficiency of the model improves with hidden size, and the batch size per expert. The computation efficiency of the model depends on the size of the underlying matrix multiplications (MM) in the model with larger MM resulting in better efficiency. The size of this MM is directly proportional to the hidden size and batch size per expert, which justifies the results we observe.

It is important to note that the batch size per expert is a function of the global batch size, number of experts, top-1 vs top-2 gating, so it is necessary to keep these parameters in mind when designing an MoE architecture and choosing a batch size for training. 

\begin{table}[htbp]
\begin{center}
{
\begin{tabular}{ll} \toprule 
Hidden Dimension & TFLOPS per GPU \\ \midrule
2048 & 75 \\
2560 & 89 \\ 
3072 & 102 \\
4096 & 120 \\
\bottomrule
\end{tabular}
}
\end{center}
\caption{Impact of Hidden Dimension on Computation Efficiency.}
\label{tab:flops} 
\end{table}

\begin{figure}[htbp]
 \centering
\includegraphics[width=1.0\linewidth]{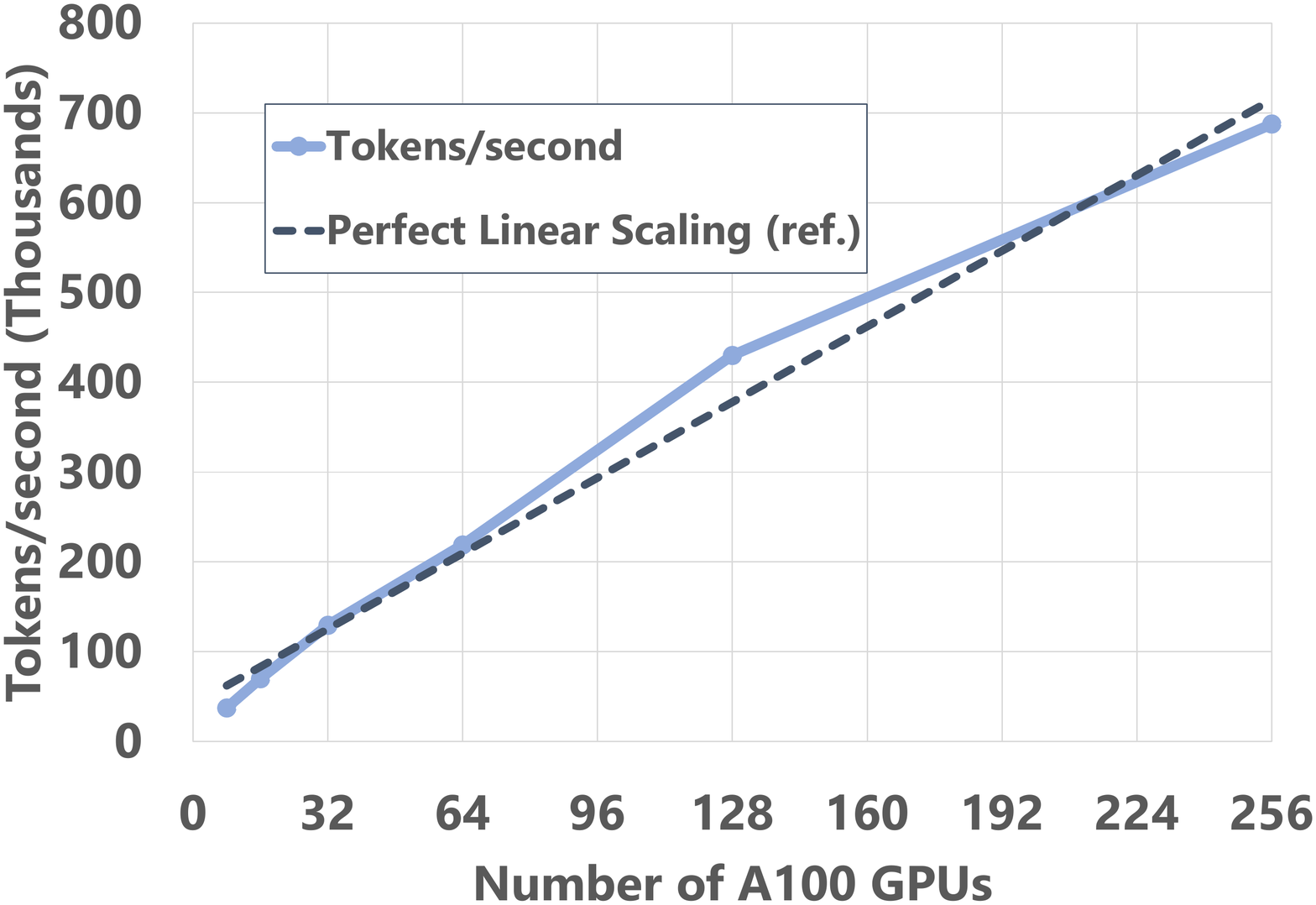}
\caption{DeepSpeed MoE throughput scaling.}
\label{fig:dsthroughput}
\end{figure}

\textbf{Scalability} In addition to excellent computational efficiency, DeepSpeed MoE also offers near perfect scalability resulting from highly optimized expert parallelism and ZeRO implementations in DeepSpeed and careful use of efficient communication primitives to minimize the communication overhead. As shown in Figure~\ref{fig:dsthroughput}, we observed near-linear increase in throughput as the number of GPUs increase. 

\subsection{DeepSpeed MoE API}

DeepSpeed MoE API is simple and easy-to-use. It has two main interfaces: 1) an MoE layer interface to extend the model and 2) a process group interface to transparently utilize multiple dimensions of parallelism supported by DeepSpeed MoE.

To extend an existing model with MoE layers, users need to add just two lines of codes as follows.

\begin{lstlisting}
from deepspeed.moe.layer import MoE
self.experts = MoE(hidden_size=model_dim, expert=ExpertModule(), num_experts="desired number of experts")
\end{lstlisting}

In addition to the MoE layer, users need to initialize the groups interface so that DeepSpeed can orchestrate communication for different parallelism dimensions without any user involvement.

\begin{lstlisting}
deepspeed.utils.groups.initialize(ep_size="desired expert-parallel world size")
\end{lstlisting}

A self-contained example to use this API can be seen from the DeepSpeed website.  \footnote{\url{https://www.deepspeed.ai/tutorials/mixture-of-experts}}

\section{Efficient MoE training and inference methods}
\label{efficient_moe}

The unique architecture of MoE models introduces new kinds of intrinsic problems and new ways of model training and inference at the same time. Here, we design and implement Random Token Selection to address the challenge of token selection bias caused by MoE model architecture and capacity limits. We exploit the autonomy of individual experts to achieve faster model convergence and better runtime efficiency.

\subsection{Random Token Selection for better and faster convergence}
\label{sec:rts}

MoE model training has a biased selection problem towards the prefix of the input sequences. The problem arises from each expert having limited capacity and the routing not being perfectly balanced among the experts. This results in less training tokens from the suffix of the sequence successfully being routed to one of the experts. When a certain expert gets selected by more tokens than its capacity, the routing algorithm usually drops the tokens at the end as \textit{overflown} tokens. This results in biasing  towards the prefix of the sequences and less training signal from the suffix of the sequence. This problem is worsened when a batch is flattened along the batch and sequence length dimensions, this happens before the All-to-All collective operation, which requires the tensor shapes to be the same across the different GPUs performing the operation. When a batch is flattened, the skew toward the front tokens happens on a sentence level. Depending on the input batch, it's possible that whole sentences may fail to have assigned expert capacity at a given update. A possible mitigation is to have an additional tensor dimension called group dimension which is used to batch multiple sentences in a bucket \cite{lepikhin2020gshard}. The selection then happens inside a given group and therefore deals with the bias towards the beginning of the sentences, though the biased selection toward the beginning of the group still remains. 

To address this issue, we introduce a simple, yet efficient  method called ``Random Token Selection (RTS)''. We randomize the priority of tokens in the flattened tensor for expert capacity assignment. By doing this, a token's chance of occupying the capacity of its selected expert becomes unrelated to its position, and only depends on the popularity of that expert among all tokens. We observe this not only removes the biased selection problem, but also works as a good regularizing technique for the large scale MoE models. This eventually results in faster convergence and makes the MoE training more efficient.

\subsection{Aggregation of experts for faster convergence}
\label{sec:aoe}
The MoE architecture naturally brings an interesting opportunity to efficiently increase model size, thanks to its conditional computation.
Expert layers are conditionally activated at a time, and they learn some expertise during the training. By collecting those expertise together, we can create a larger mixture of experts, which we call ``Aggregation of Experts (AoE)''. For the actual implementation, we combine widely used checkpoint averaging technique with concatenation of experts and gating matrices. First, two checkpoints are selected to be aggregated. Then, all the weights except for the experts and gating layer are averaged. Lastly, the gating layer matrices are concatenated and expert weights from two checkpoints are collected to one checkpoint. After this procedure, this newly created checkpoint can be used as an initializer for a new model and the model can be continuously trained or fine-tuned towards a downstream task.

\subsection{Experts pruning for faster inference}
With a similar intuition to the aggregation of experts approach, experts can be pruned to compress model for a better runtime efficiency.
During training, the model tries to utilize the large scale parameters to learn a generalized representation. At the same time, the sparse experts can get specialized into particular aspects of the tasks. Therefore, we hypothesize that a subset of experts may be capable of handling the learned tasks to a certain extent.

To explore this opportunity, we experiment with a pruning technique, where the network is kept as it is except by selecting a subset of the expert weights, FFNs and gates. We try two methods of selecting the experts: random selection and a utilization-based approach where we select the $k$ experts with the highest utilization. For the utilization-based approach, we simply count the number of times a token is routed to each expert in the validation set. This simple scheme can be used to easily extract smaller checkpoints as needed for practical deployments of these models for downstream tasks.

\section{Multitask Multilingual MoE:  Z-code M3}
\label{sec:model_train}

\begin{table}[]
\begin{center}
\begin{tabular}{lr}
\toprule
\textbf{Model} & \textbf{Parameters} \\ \midrule
Z-code          & 0.7B               \\ 
Z-code M3 8 experts      & 1.8B                 \\ 
Z-code M3 16 experts     & 3B                 \\ 
Z-code M3 32 experts     & 5.5B                 \\ 
Z-code M3 64 experts     & 10B                 \\ 
Z-code M3 128 experts     & 20B                 \\ 
\bottomrule
\end{tabular}
\end{center}
\caption{Comparison of model sizes.}
\label{tab:model-sizes}
\end{table}

\subsection{Multitask multilingual training}
\label{sec:multitask}
Multitask training framework has shown effectiveness for improving the performance of various language tasks. Here, we exploit multitask training as a method to further improve MoE models' quality. Based on \cite{wang-etal-2020-multi} and ELECTRA\cite{clark2020electra}, we experiment with combinations of various tasks including Machine Translation (MT),  Denoising Auto-Encoding (DAE), ELECTRA , noised MT,  and Masked Language Model(MLM). While supervised tasks such as MT would suffer from data limitation, the rests  are  unsupervised tasks that do not have such limitation. This enables us to explore various tasks that can utilize larger models and learn better language representation. In this work,  we leverage the scalability of MoE models to improve  pretraining of multitask and multilingual models and better help downstream tasks through utilizing the larger model capacity provided by MoE architecture. 

We use MT as the main task which we use to track the validation loss. This is similar to GShard model \cite{lepikhin2020gshard} which was trained to perform  MT task only. Based on the experiments, we add Denoising  Auto-Encoding (DAE) task similar to \cite{maurya-etal-2021-zmbart} and \cite{xue-etal-2021-mt5} which are targeting generation tasks as a co-training task. For the DAE task, we follow the settings from \cite{wang-etal-2020-multi}. We use the target-side monolingual data for DAE task training. We utilize three types of noises  which are text infilling, word drop \& word blank and word swapping. In this task, monolingual data is used to generate noisy target data with the task is to restore the original sequence. Furthermore, we utilizes Masked Language model similar to \cite{devlin2018bert} and experiment with  ELECTRA \cite{clark2020electra}  using monolingual data which are  applied to the encoder part of the model only. 

One thing to note about the reference models mentioned above, a model is usually trained on a single task. For example, Gshard  \cite{lepikhin2020gshard} with up to 600B parameters model was trained on MT task only. While Switch \cite{fedus2021switch} trained on DAE task and so on. Our model called, Z-code M3 (Multitask, Multilingual and MoE) model tries to harvest the combined benefits of such diverse tasks through the  scalablity  of MoE architecture.

At training time we sample a batch per task; effectively the model would consume the same numbers of batches from each task regardless of the data sizes per task. Within the same task, the multilingual data distribution is usually skewed towards high-resource languages. We utilize temperature sampling as detailed in \cite{wang-etal-2020-multi} to balance data sampling across different languages. We use simple additive loss with all tasks. For instance, when training MT and DAE  objectives together, we add the losses as below:

\begin{equation*}
\mathcal{L} = \mathcal{L}_{MT} + \mathcal{L}_{DAE}
\end{equation*}

\subsection{Model details}
We use transformer encoder-decoder architecture\cite{vaswani2017attention} as the base model. For better inference efficiency, we use deeper encoders and shallower decoders architecture presented in \cite{kim-etal-2019-research} and \cite{kasai2020deep}. We use smaller model architecture for the small scale ablation studies,  which has 12 encoder layers and 6 decoder layers. Each transformer block has 768 hidden embedding dimension and 3072 feed-forward layer hidden dimension with 12 multi-head attention heads. For the large scale multitask training setting, we use 24 encoder layers and 12 decoder layers with 1024 hidden dimension and 4096 feed-forward layer hidden dimension with 16 multi-head attention heads. We use pre-layer normalization which is becoming more standard for the transformer architecture presented in \cite{xiong2020layer}. Regarding  MoE layer placement, we use an expert layer every other layer following \cite{lepikhin2020gshard} and \citep{fedus2021switch}. We use  250,000 size vocabulary with sentencepiece\footnote{\url{https://github.com/google/sentencepiece}} library. The base model size is 0.7B parameters,  Table \ref{tab:model-sizes} shows the number of parameters corresponding to different number of experts.

\section{Experiments}
\label{sec:exp}
\subsection{Experimental setup}
\label{subsec:exp-setup}
\paragraph{Task}
During model pretraining, we report the MT cross-entropy objective loss as the validation loss metric.
\paragraph{Data}
We  start by evaluating  the models on a small scale machine translation single task training on WMT-10 benchmark dataset \cite{wang-etal-2020-multi}. This dataset is a collection of parallel data in different languages from the WMT
shared tasks. Each corpus is composed of sentences in English
(En) and in other 10 languages: French (Fr),
Czech (Cs), German (De), Finnish (Fi), Latvian
(Lv), Estonian (Et), Romanian (Ro), Hindi (Hi),
Turkish (Tr) and Gujarati (Gu). There is a total of 32.5 million sentence pairs in the training set. We combine
all the parallel corpus together as the
training set and evaluate the models on the test sets
from each language. We report the case-sensitive detokenized BLEU using sacreBLEU \footnote{\url{https://github.com/mjpost/sacreBLEU}}. 

For the large scale and multitask training settings, we utilize a dataset composed of 336 billion tokens in 50 different languages. It is composed of an in-house web-crawled parallel dataset with 43 million tokens joined together with 293 billion monolingual tokens from CCNet\cite{wenzek-etal-2020-ccnet} covering 50 languages. More information on the data sizes can be found in the appendix.
We evaluate the translation task on various testsets from WMT as described in \cite{wang-etal-2020-multi} and on an in-house parallel testset composed of 98 language directions. 
\paragraph{Hardware}
Models are trained on 40GB A100 GPUs. The number of GPUs used vary according to the size of the models.
\paragraph{Hyperparameters}
For training, we use the Adam optimizer with 5000 warm-up steps, a start learning rate of 0.03, and inverse square root scheduler as proposed on \cite{raffel2020exploring}. We accumulate gradients to make an effective batch size of 1.5 million tokens per task. For the DAE task, we use an infill ratio of 0.2, and dropout/blank probability of 0.1.
\paragraph{MoE setting}
We explore the hyperparameters for the MoE layers and use the setting that performs the best. We observe top-1 gating from Switch Transformers \cite{fedus2021switch} is more effective than the top-2 gating from GShard \cite{lepikhin2020gshard}. Similar to the setup in Switch models \cite{fedus2021switch}, we use capacity factor of 1.0 for the training and 2.0 for the evaluation. For a better exploration of new experts and regularization of the model, we apply jittering noise on the gating input following Switch transformers \cite{fedus2021switch} and also apply the same truncated normal distribution initialization for all weight matrices. Finally, we use an additional balancing loss together with the cross-entropy loss to better balance the utilization of different experts. The balancing loss is multiplied by 0.01 and added to the total loss.

\subsection{MoE sample efficiency}
We evaluate the model training with different number of experts. We train five models with different number of experts settings: 1, 8, 16, 32 and 64 experts. Except for the number of experts, all  other model settings are kept the same. The model setting with one expert does not have gating and routing layers, so it is a plain dense Z-code architecture. The models with more than one expert are trained up to 20,000 update steps. In order to compare the steps to reach cross-entropy loss value of 4.3, the Z-code [0.7B] dense model, which is equivalent in size to single expert model, needs to be trained with much more steps.

\begin{figure}[!ht]
  \centering
  \includegraphics[width=0.9\linewidth]{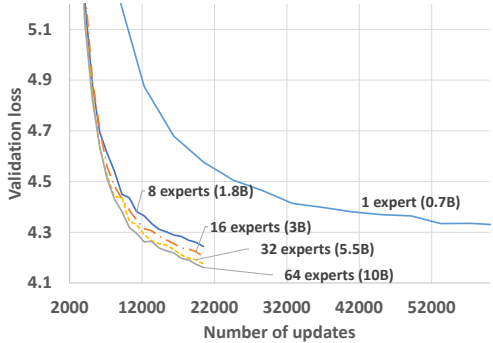}
\caption{Better Model training with MoE.}
\label{fig:capacity}
\end{figure}

Figure \ref{fig:capacity} shows the validation loss curves of those training experiments. Any model setting that has more than one expert outperforms the Z-code [0.7B] model. The model with 64 experts can reach the same loss value with ten times less update steps compared to Z-code [0.7B] model. This shows that MoE model is much more sample efficient than the plain dense model. It is also observed that the model converges with less updates when the number of experts increases. It is worth noting that there is a diminishing return with the number of experts. 
We observe that models with fewer experts do similarly well at the beginning, but the models with more experts start to outperform when the training steps progress.

\begin{figure}[!ht]
  \centering
  \includegraphics[width=0.9\linewidth]{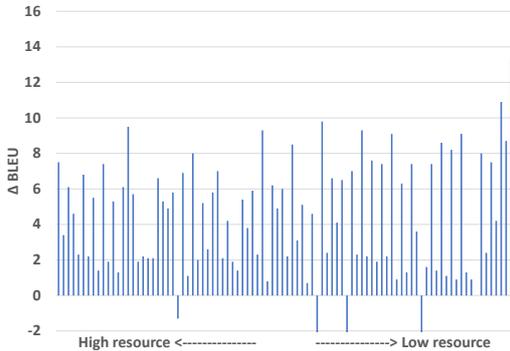}
\caption{MT BLEU score improvements with MoE and more experts.}
\label{fig:mtbleu}
\end{figure}

Figure \ref{fig:mtbleu} shows the final BLEU score gains of Z-code M3 [10B] model by adding 64 experts compared to the Z-code [0.7B] model. The chart includes 98 from and to English language pairs. In most cases, Z-code M3 model outperforms Z-code [0.7B] model significantly. The gain is higher for the low resource languages, but consistent gains are observed even in high resource languages. This shows that MoE models can utilize various parameters together to handle diverse problems effectively.

\begin{table}[!ht]
\begin{center}\small
{
\scriptsize
\begin{tabular}{lrrr} \toprule 
 & X to English & English to X & Average \\ \midrule
Non-MoE multilingual model & 35.16 & 30.36 & 32.76 \\
Individual bilingual models & 35.70 & 30.40 & 33.05 \\
Z-code M3 [10B] & 42.09 & 32.22 & 37.15 \\
\bottomrule
\end{tabular}
}
\end{center}
\caption{BLEU score comparison between different models for an in-house 50 language test dataset.}
\label{tab:mt-gain}
\end{table}

\begin{table}[!ht]
\begin{center}\small
{
\scriptsize
\begin{tabular}{lrrr} \toprule 
 & X to English & English to X & Average \\ \midrule
M2M-100 [12B] & 33.70 & 30.08 & 31.89 \\
Z-code M3 [10B] & 37.53 & 32.17 & 34.85 \\
\bottomrule
\end{tabular}
}
\end{center}
\caption{BLEU score comparison between different models for the public WMT test dataset from/to English (En) to/from (French (Fr),
Czech (Cs), German (De), Finnish (Fi), Latvian
(Lv), Estonian (Et), Romanian (Ro), Hindi (Hi),
Turkish (Tr)). Language pairs including Gu are removed because M2M system does not support Gu.}
\label{tab:mt-gain}
\end{table}

\subsection{Multitask training with MoE}
We evaluate the effectiveness of MoE model architecture with the multitask training setting. As explained in Section \ref{sec:multitask}, we use MT and DAE tasks together. For comparing the performance of multitask training, we use MT validation loss value which is commonly available in both MT task only and multitask settings. In this experiment, we use plain Z-code model which has 700 million parameters and Z-code M3 model with 64 experts which has 10 billion trainable parameters.

\begin{figure}[!ht]
  \centering
  \includegraphics[width=1.0\linewidth]{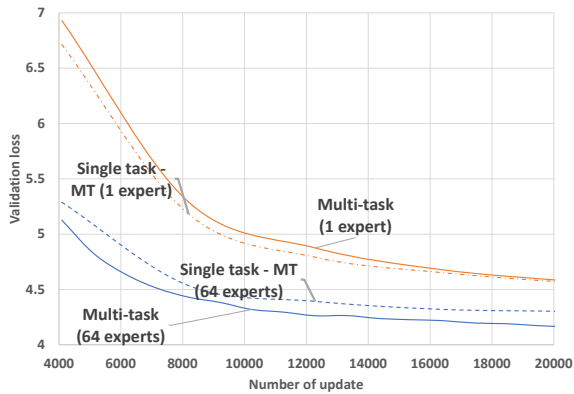}
\caption{Better Multitask training with MoE.}
\label{fig:multitask}
\end{figure}

Figure \ref{fig:multitask} shows how two different models converges with single task and multitask settings. The Z-code [0.7B] model converges faster with only the MT task and the final loss does not improve with a multitask setting. On the other hand, Z-code M3 model shows significantly faster convergence with multitask training setting compared to MT only training. The final converged loss value is also better with multitask training.

\begin{figure}[!ht]
  \centering
  \includegraphics[width=1.0\linewidth]{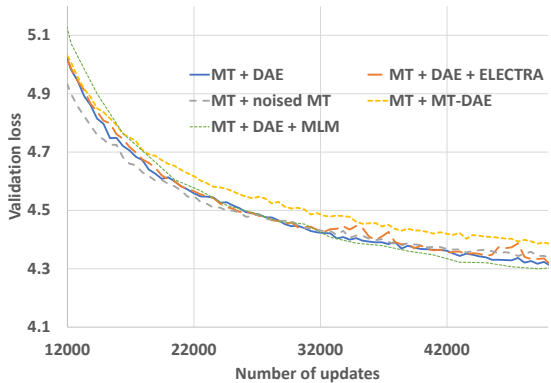}
\caption{Different Multitask objectives pretrained with 8 experts (1B parameters).}
\label{fig:ablationmultitask}
\end{figure}

Figure \ref{fig:ablationmultitask} shows additional experiments in a smaller setting (8 experts, 1 billion parameter model), with different multitask combinations as explained in \ref{sec:multitask}. In the figure, we can observe that the MT + DAE objective yields a good tradeoff in the MT validation loss. The MT + DAE + ELECTRA and MT + DAE + MLM are slightly worse and better respectively, than MT + DAE. MT + DAE is preferred for improving machine translation main objective due to its good performance and reduced complexity compared to optimizing three different tasks such as MT+DAE+MLM setting.  MLM and ELECTRA tasks are more useful for encoder-only downstream tasks, we leave it for future work to perform comprehensive  experiments to validate this on the multitask setup.

\subsection{Downstream tasks}

To further validate the performance improvements of Z-code M3 models, we also verify the 64 experts model's language generation capability on two downstream tasks. We first fine-tune a model on Wikilingua \cite{ladhak-etal-2020-wikilingua} which is a Natural Language Generation (NLG) task part of the Gem Benchmark \cite{gembenchmark2021} and show that our model significantly outperforms our baselines.
We then study multilingual non-English centric translation task (XY) and show that our model is able to improve across the board for language-pairs that were not originally part of pretraining. Finally, we fine-tune the model on the XNLI classification \cite{conneau2018xnli} task, following \cite{raffel2020exploring} we cast the task as a sequence generation problem.

As for our strategy, we first pretrain a Z-code M3 [10B] model for 170k steps, with a batch size of 1.5M tokens per task (with MT and DAE objectives) which amounts to 613B tokens. We then separately fine-tune this model on each downstream task. Following \cite{fedus2021switch}, since our MoE model routes to one expert per layer, we evaluate to an equivalent dense baseline (Z-code [0.7B] below), which matches the FLOPs computation and is the base of our MoE models.

%% Wikilingua
\paragraph{Wikilingua}
Since Wikilingua has no corresponding public leaderboard, our baseline is a carefully fine-tuned dense Z-code \cite{wang-etal-2020-multi} model and mBart \cite{liu2020multilingual}.
Table \ref{tab:wikilinguatable} shows the fine-tuning results of Z-code M3 [10B] on the Wikilingua task which we can see significant gains  compared to the baselines. We observe around 7.5 point ROUGE increase by averaging across the results from each test splits with respect to the mBART and Z-code dense baseline. This shows that our model is able to learn generic multilingual representation through  the multitask setup.
We can see a further improvement of around 10 ROUGE points increase by adding DAE as an additional task during fine-tuning and increasing the DAE task difficulty by linear scheduling, which also shows the benefits of multitask training though during  finetuning this time.

\begin{table*}[!ht]
\centering
{
\small
\begin{tabular}{|c|c|l|l|c|l|l|c|l|l|c|l|l|}
\hline
\textbf{Model}          & \multicolumn{3}{c|}{\textbf{Es-En}}    & \multicolumn{3}{c|}{\textbf{Tr-En}}    & \multicolumn{3}{c|}{\textbf{Ru-En}}    & \multicolumn{3}{c|}{\textbf{Vi-En}}    \\ \hline
\textbf{mBART}          & \multicolumn{3}{c|}{38.3/15.37/32.4}   & \multicolumn{3}{c|}{33.68/12.74/27.62} & \multicolumn{3}{c|}{32.91/11.83/27.69} & \multicolumn{3}{c|}{31.89/11.07/26.36} \\ \hline
\textbf{Z-code [0.7B]}    & \multicolumn{3}{c|}{37.78/16.38/32.09} & \multicolumn{3}{c|}{48.45/27.14/42.15} & \multicolumn{3}{c|}{37.64/16.92/32.15} & \multicolumn{3}{c|}{40.72/19.85/34.89} \\ \hline
\textbf{Z-code M3 [10B]}       & \multicolumn{3}{c|}{45.46/23.71/39.37}   & \multicolumn{3}{c|}{57.49/38.46/52.49} & \multicolumn{3}{c|}{46.52/25.75/41.13} & \multicolumn{3}{c|}{49.31/28.78/44.06} \\ \hline
\textbf{Z-code M3[10B] + DAE} & \multicolumn{3}{c|}{\textbf{47.12/29.39/42.38}}  & \multicolumn{3}{c|}{\textbf{66.88/55.24/64.24}} & \multicolumn{3}{c|}{\textbf{46.87/30.28/42.69}} & \multicolumn{3}{c|}{\textbf{53.86/38.97/50.27}} \\ \hline
\end{tabular}
\caption{The mBART model has 0.68B parameters. The Z-code dense model has 0.7B, and the Z-code M3 models 10B. The Z-code M3 + DAE model, is fine-tuned in a multitask fashion with the DAE objective loss besides MT objective, and the DAE parameters are scheduled linearly following \cite{wang-etal-2020-multi}. Each row describes the ROUGE-1/ROUGE-2/ROUGE-L F1 scores respectively, as reported in \cite{ladhak-wiki-2020}.}
\label{tab:wikilinguatable}
}
\end{table*}

%% XY dataset
\paragraph{XY translation}  We experiment with multilingual non-English centric translation task (XY). We use a test dataset composed of 704 language pairs. Each language-pair corpus contains 1000 sentences where the source data-side data comes from the WMT dataset and the target-side data is human translated.
For fine-tuning, we use a subset of the CCMatrix dataset  \cite{schwenk-etal-2021-ccmatrix} composed of 241M parallel sentences spanning 1114 language pairs. We follow the same pretraining strategy previously described but use a smaller batch size of 450k tokens. We also employ \textit{expert dropout} of 0.4, as described in \cite{fedus2021switch} since this significantly improves the model's final performance. As for the baselines, we also compare against the dense Z-code [0.7B] fine-tuned model on the same data and on the publicly available M2M-100 1.2B model \cite{fan2020fbm2m}.

Table \ref{tab:xy-finetuning} summarizes our results. We observe an average BLEU increase of 1.37 when compared with our baseline and a BLEU increase of 1.02 when compared to the M2M-100 [1.2B] model. This shows that our model, in comparison to its dense counterpart, is able to better generalize to non-English centric language pairs not seen in pretraining and is therefore an effective multi-lingual learner.

\begin{table}[!ht]
\centering
{
\small
\begin{tabular}{lrr} \toprule
\textbf{Model}                 &  \textbf{BLEU} &  \textbf{BLEU $\Delta$}  \\ \midrule
Z-code [0.7B] (ref.) $\dagger$ & 20.34     &  -                                  \\
M2M-100 [1.2B]               & 20.69    &    0.35                                      \\
Z-code M3 [10B]$\dagger$   & \textbf{21.71}    &    1.37                                      \\
\bottomrule
\end{tabular}

\caption{Average BLEU scores comparison for non-English centric fine-tuned multilingual models and evaluated on a XY testset composed of 704 language pairs. The BLEU $\Delta$ refers to the score difference to the $1^{st}$ row (reference) and the $\dagger$ signifies that the model was fine-tuned. }
\label{tab:xy-finetuning}
}
\end{table}

%% XNLI results

\paragraph{XNLI} To validate the capabilities of Z-code M3 model on various tasks, we evaluate the model on XNLI entailment task\cite{conneau2018xnli} covering 14 languages. Table \ref{tab:xnli-results} shows the results on the XNLI dataset. The model is trained as a generation task, similar to MT and summarization tasks.  We compare  to an equivalent dense baseline (Z-code [0.7B] below), which matches the FLOPs computation and is the base of our MoE models. We also compare to mT5 base model (580M parameters) \cite{xue-etal-2021-mt5} as a equivalent dense model trained on single task though on much larger training data than our model. We can observe Z-code M3 model is performing much  better than Z-code [0.7B] dense model as well as mT5 base [0.6B] on both zero-shot and translate train scenarios. We can see a clear advantage of the sparse model compared to its dense counterpart, there is approximately 3 accuracy points gain across both zero-shot and translate-train settings.

\begin{table}[!ht]
\centering
{
\small
\begin{tabular}{cc} \toprule
% \hline
\textbf{Model}            & \textbf{Accuracy}        \\ \midrule %\hline
\multicolumn{2}{c}{{\color[HTML]{000000} \textit{Zero-shot}}}       \\ \midrule %\hline
XLM-R & 79.2 \\
mT5 base [0.6B]              & 75.4                   \\
Z-code [0.7B]               & 77.42                    \\ %\hline
Z-code M3 [10B]                  &  \textbf{80.27}                    \\ \midrule %\hline
\multicolumn{2}{c}{{\color[HTML]{333333} \textit{Translate-train}}} \\ \midrule %\hline
XLM-R & 82.6 \\
mT5 base [0.6B]              & 75.9                   \\
Z-code [0.7B]               & 79.59                    \\ %\hline
Z-code M3 [10B]                  & \textbf{82.77}                    \\ \bottomrule %\hline
\end{tabular}
\caption{XNLI test accuracy results. The zero-shot setting trains the model with the original English only dataset and evaluate on the target languages. The translate-train setting augments the dataset by translating the original examples on each target language.}
\label{tab:xnli-results}
}
\end{table}

\subsection{RTS ablation study}
We conduct an ablation study for the performance of three token assignment algorithms described in Section \ref{sec:rts} on the WMT dataset. First, a plain approach that uses all tokens across different sentences into one bucket. Second, the algorithm where the token selection is done in different groups separately. Lastly, the random token selection algorithm that randomly selects the priority of tokens inside a bucket.

\begin{figure}[!ht]
  \centering
  \includegraphics[width=0.9\linewidth]{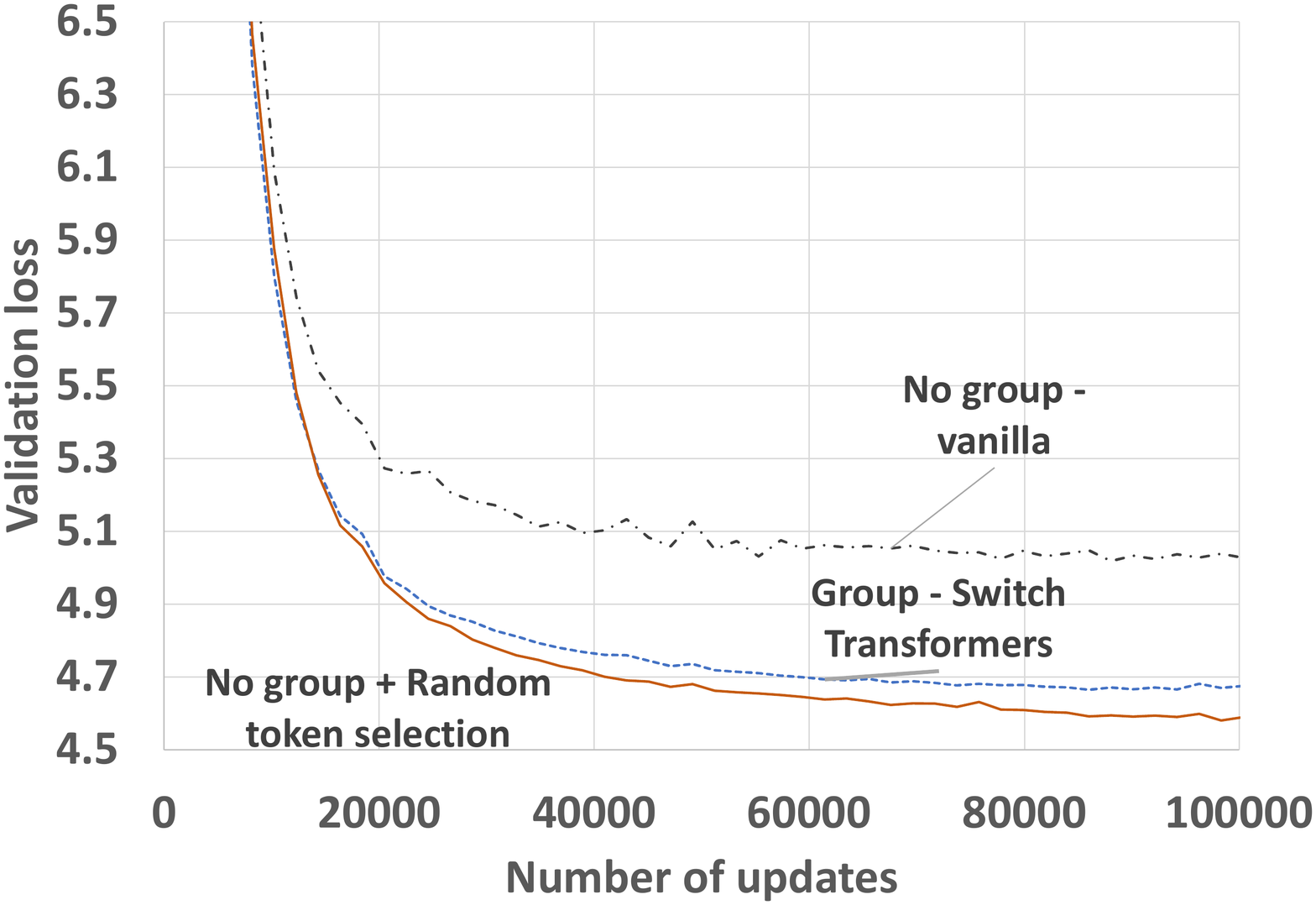}
\caption{Better convergence with random token selection.}
\label{fig:rts}
\end{figure}

In the Figure \ref{fig:rts}, the convergence curves of those three algorithms are compared. The plain algorithm converges slowly and suffers severely from the biased token selection. However, by adding the grouping, the convergence improved quite much. One thing to note is that the grouping approach makes the batch size harder to be dynamically changed. This prevents the efficient utilization of training batches by having redundant padding tokens.

On the other hand, random token selection method without introducing a group dimension show even better convergence than the grouping approach. We hypothesize that it removes token selection bias and regularize the model training more by dropping tokens in a random way. This method does not prevent dynamic batching which yields better throughput.

\subsection{AoE experiments}
In this experiment, we pick two checkpoints with 64 experts each and merge them into a single checkpoint following the method we present in section \ref{sec:aoe}. This results in a single checkpoint with 128 experts which is then used to initialize a model with 128 experts that is then trained. As a comparison study, we also train an equivalent same-sized model with 128 experts but with the standard initialization scheme described in Section \ref{subsec:exp-setup}.

\begin{figure}[!ht]
 \centering
\includegraphics[width=1.0\linewidth]{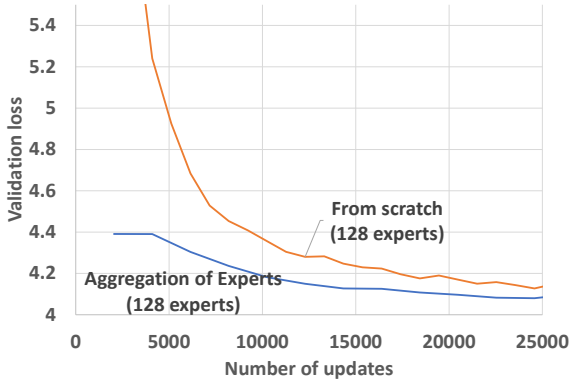}
\caption{Validation loss curve comparison between aggregation of experts initialized training and random initialized training.}
\label{fig:aggexperts}
\end{figure}

In Figure \ref{fig:aggexperts}, the top loss curve represents the training of a randomly initialized model (baseline) and the curve below is the training of the model initialized with the merged checkpoint. We can clearly see that the model initialized with our proposed technique converges much faster than the baseline.

\subsection{Experts pruning experiments}
\label{top-exps-experiments}

\begin{figure}[!ht]
  \centering
  \includegraphics[width=1.0\linewidth]{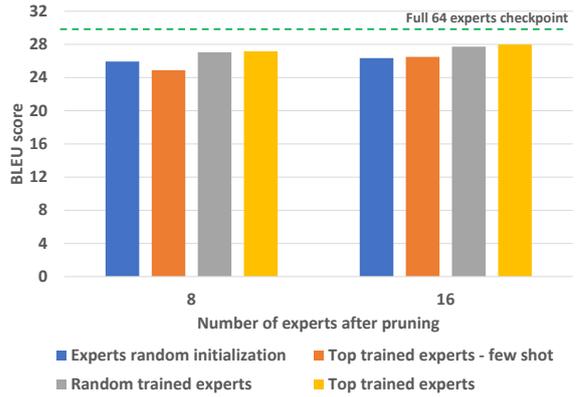}
\caption{BLEU score comparison between model trained from scratch and the models with pruned and fine-tuned. All experiments were trained for 10k updates.}
\label{fig:pruning}
\end{figure}

Figure \ref{fig:pruning} shows the results of extracting 8 and 16 experts out of a checkpoint fully trained with 64 experts, on the MT and DAE tasks. The experts are then fine-tuned on the same tasks, and compared to baselines trained from scratch with the same parameter count for a small number of updates. It is worth to note that using extracted experts converges much faster than training from scratch, and retains a competitive BLEU score compared to the full checkpoint. In fact, we observe that at the first BLEU evaluation, at the 640th update, the pruned checkpoints achieve an already quite close BLEU score to the checkpoints trained more than 10K updates with random initialization. Looking at the weights cosine distance, these early updates focus on adapting the gating layer weights, which shows the benefits of a modular model where experts are trained to specialize at some extent. We also observe that evaluating the pruned checkpoints without any fine-tuning at all do not yield very good results, however it is far from being completely random, which reinforces the need to adapt special weights on the early updates, such as the gating layer. Finally, we observe the random experts perform slightly worse than the top experts, which shows empirically that the experts are uniformly well trained and work well independently.

\section{Conclusion}
In this work, we have introduced an efficient way of scaling MoE models with DeepSpeed MoE and have shown the effectiveness of jointly applying the sparse MoE architecture and multitask training on different multilingual settings while employing the machine translation training objective. Our proposed methods not only achieve state-of-the-art translation and NLG results, but also significantly improve training efficiency. 
We propose RTS (Random Token Selection) to resolve the intrinsic expert capacity issue and achieve a better sample efficiency. Also, by exploiting the sparse MoE architecture, we propose two flexible methods which are AoE (Aggregation of Experts) and Experts pruning for scaling up and compressing such models to enable more efficient training and inference on a wider range of scenarios.
Finally, we confirm once again the importance of efficient methods and software tooling to scale up MoE models. 

As for future work, we are looking at more efficient integration  of ZeRO-Offload with Z-code M3 models to allow very large base model sizes. We plan to examine the limits and returns of scaling MoE models while efficiently utilizing the computation resources. We are also studying methods to improve the utilization of experts and routing algorithms for specific tasks and/or languages, which we found to be a limiting factor for improving translation from English into other languages.

\clearpage

\bibliographystyle{acl_natbib}
\bibliography{moe-arxiv}

\onecolumn
\appendix
\section*{Appendix}
\section{Languages and monolingual data statistics}

\begin{table*}[!h]
\centering
\begin{tabular}{c l r r | c l r r}
\hline
\textbf{ISO} & \textbf{Language} & \textbf{Tokens} & \textbf{Size} & \textbf{ISO} & \textbf{Language} & \textbf{Tokens} & \textbf{Size} \\
\textbf{code} &  & (M) & (GiB) & \textbf{code} & & (M) & (GiB) \\ \hline
% \textbf{ISO code} & \textbf{Language} & \textbf{Tokens} (M) & \textbf{Size} (GiB) & \textbf{ISO code} & \textbf{Language} & \textbf{Tokens} (M) & \textbf{Size} (GiB) \\ \hline
af & Afrikaans & 262 & 1.4 & ko & Korean & 1138 & 11.3 \\
ar & Arabic & 2869 & 28.0 & lt & Lithuanian & 1903 & 14.2 \\
bg & Bulgarian & 5649 & 59.2 & lv & Latvian & 1264 & 9.3 \\
bn & Bengali & 988 & 15.7 & ml & Malayalam & 627 & 15.6 \\
cs & Czech & 2648 & 17.3 & mr & Marathi & 438 & 7.4 \\
da & Danish & 7900 & 46.0 & ms & Malay & 71 & 0.48 \\
de & German & 10297 & 66.5 & my & Burmese & 15 & 0.44 \\
el & Greek & 4289 & 46.9 & nl & Dutch & 5207 & 30.4 \\
en & English & 55647 & 300.8 & pl & Polish & 6786 & 46.6 \\
es & Spanish & 9374 & 53.2 & pt & Portuguese & 8405 & 49.1 \\
et & Estonian & 843 & 6.1 & ro & Romanian & 10367 & 61.5 \\
eu & Basque & 270 & 2.0 & ru & Russian & 23408 & 277.9 \\
fa & Persian & 13579 & 114.3 & sk & Slovak & 3592 & 23.7 \\
fi & Finnish & 6839 & 55.2 & sl & Slovenian & 1669 & 10.3 \\
fr & French & 9780 & 56.7 & sv & Swedish & 13210 & 79.3 \\
ga & Irish & 188 & 1.1 & sw & Swahili & 275 & 1.6 \\
gl & Galician & 495 & 2.9 & ta & Tamil & 1200 & 25.7 \\
he & Hebrew & 3513 & 32.7 & te & Telugu & 660 & 12.8 \\
hi & Hindi & 1767 & 20.9 & th & Thai & 1834 & 71.6 \\
hr & Croatian & 3445 & 21.4 & tl & Tagalog & 556 & 3.1 \\
hu & Hungarian & 7828 & 58.5 & tr & Turkish & 3427 & 26.5 \\
id & Indonesian & 23246 & 151.8 & ur & Urdu & 748 & 5.8 \\
it & Italian & 7535 & 45.6 & vi & Vietnamese & 25779 & 143.0 \\
ja & Japanese & 530 & 69.2 & yo & Yoruba & 0.86 & 0.01 \\
ka & Georgian & 469 & 9.1 & zh & Chinese & 259 & 46.9 \\
\hline
\end{tabular}
\caption{Languages and monolingual data statistics used from CCNet in multi-task training. We report the list of 50 languages and the number of tokens (Millions) and the size of the data (in GiB) for each language.}
\label{tab:monolinugal-appendix-stats}
\end{table*}

\clearpage
\section{Parallel data statistics}

\begin{table*}[!h]
\centering
\begin{tabular}{c c c c c | c c c c c}
\hline
\multirow{2}{*}{\textbf{Language pair}} & \multicolumn{2}{c}{\textbf{Tokens} (M)} & \multicolumn{2}{c}{\textbf{Size} (GiB)} & \multirow{2}{*}{\textbf{Language pair}} & \multicolumn{2}{c}{\textbf{Tokens} (M)} & \multicolumn{2}{c}{\textbf{Size} (GiB)} \\
\cmidrule(rl){2-3} \cmidrule(rl){4-5} \cmidrule(rl){7-8} \cmidrule(rl){9-10} 
& \textbf{non-En} & \textbf{En} & \textbf{non-En} & \textbf{En} & & \textbf{non-En} & \textbf{En} & \textbf{non-En} & \textbf{En} \\
\hline
$Af \leftrightarrow En$ & 23 & 22 & 0.13 & 0.13 & $Lt \leftrightarrow En$ & 124 & 157 & 0.94 & 0.91 \\
$Ar \leftrightarrow En$ & 410 & 453 & 4.0 & 2.6 & $Lv \leftrightarrow En$ & 136 & 169 & 1.0 & 0.97 \\
$Bg \leftrightarrow En$ & 306 & 313 & 3.2 & 1.7 & $Ml \leftrightarrow En$ & 18 & 27 & 0.44 & 0.14\\
$Bn \leftrightarrow En$ & 28 & 31 & 0.43 & 0.16 & $Mr \leftrightarrow En$ & 20 & 23 & 0.34 & 0.13\\
$Cs \leftrightarrow En$ & 525 & 615 & 3.5 & 3.4 & $Ms \leftrightarrow En$ & 16 & 17 & 0.11 & 0.10 \\
$Da \leftrightarrow En$ & 382 & 406 & 2.3 & 2.2 & $My \leftrightarrow En$ & 0.5 & 0.9 & 0.01 & 0.01 \\
$De \leftrightarrow En$ & 492 & 529 & 3.5 & 3.1 & $Nl \leftrightarrow En$ & 571 & 572 & 3.3 & 3.1 \\
$El \leftrightarrow En$ & 556 & 566 & 6.1 & 3.1 & $Pl \leftrightarrow En$ & 333 & 381 & 2.4 & 2.2 \\
$Es \leftrightarrow En$ & 447 & 414 & 2.6 & 2.3 & $Pt \leftrightarrow En$ & 144 & 151 & 0.81 & 0.78 \\
$Et \leftrightarrow En$ & 387 & 495 & 2.8 & 2.8 & $Ro \leftrightarrow En$ & 396 & 396 & 2.3 & 2.1 \\
$Eu \leftrightarrow En$ & 7.3 & 9.5 & 0.05 & 0.05 & $Ru \leftrightarrow En$ & 344 & 395 & 4.1 & 2.1 \\
$Fa \leftrightarrow En$ & 139 & 119 & 1.1 & 0.66 & $Sk \leftrightarrow En$ & 159 & 188 & 1.1 & 1.1 \\
$Fi \leftrightarrow En$ & 236 & 347 & 1.9 & 1.9 & $Sl \leftrightarrow En$ & 207 & 250 & 1.3 & 1.4 \\
$Fr \leftrightarrow En$ & 517 & 464 & 3.1 & 2.6 & $Sv \leftrightarrow En$ & 317 & 352 & 2.0 & 1.9 \\
$Ga \leftrightarrow En$ & 33 & 30 & 0.20 & 0.18 & $Sw \leftrightarrow En$ & 29 & 31 & 0.18 & 0.18 \\
$Gl \leftrightarrow En$ & 3.8 & 4 & 0.02 & 0.02 & $Ta \leftrightarrow En$ & 82 & 102 & 1.8 & 0.58 \\
$He \leftrightarrow En$ & 168 & 208 & 1.5 & 1.1 & $Te \leftrightarrow En$ & 30 & 39 & 0.60 & 0.22 \\
$Hi \leftrightarrow En$ & 274 & 242 & 3.2 & 1.3 & $Th \leftrightarrow En$ & 81 & 230 & 3.1 & 1.3 \\
$Hr \leftrightarrow En$ & 199 & 227 & 1.2 & 1.2 & $Tl \leftrightarrow En$ & 0.7 & 0.7 & 0.01 & 0.01 \\
$Hu \leftrightarrow En$ & 370 & 456 & 2.7 & 2.5 & $Tr \leftrightarrow En$ & 238 & 316 & 1.8 & 1.7 \\
$Id \leftrightarrow En$ & 52 & 57 & 0.34 & 0.30 & $Ur \leftrightarrow En$ & 27 & 22 & 0.19 & 0.11 \\
$It \leftrightarrow En$ & 261 & 254 & 1.6 & 1.4 & $Vi \leftrightarrow En$ & 203 & 151 & 1.1 & 0.82 \\
$Ja \leftrightarrow En$ & 58 & 504 & 3.6 & 2.8 & $Yo \leftrightarrow En$ & 3.1 & 2.7 & 0.02 & 0.02 \\
$Ka \leftrightarrow En$ & 17 & 21 & 0.29 & 0.12 & $Zh \leftrightarrow En$ & 39 & 481 & 2.3 & 2.8 \\
$Ko \leftrightarrow En$ & 307 & 421 & 2.8 & 2.3 & \\
\hline

\end{tabular}
\caption{In-house web crawled parallel data statistics used in multi-task training. We report the list of 98 language directions and the number of tokens (Millions) and the size of the data (in GiB) for non-English and English of each language pair.}
\label{tab:parallel-appendix-stats}
\end{table*}

\end{document}